%% file: root.tex
\documentclass[letterpaper, 10 pt, journal, twoside]{IEEEtran}

\usepackage{graphics} 
\usepackage{epsfig} 
\usepackage{mathptmx} 
\usepackage{times} 
\usepackage{amsmath} 
\usepackage{amssymb}  
\usepackage{amsmath,amssymb,amsfonts,bm}
\usepackage{booktabs}
\usepackage{multirow}
\usepackage{algorithm}
\usepackage{algorithmic}

\usepackage{soul,color}
\usepackage{hyperref}
\usepackage{cuted}
\usepackage{capt-of}
\usepackage{cleveref}
\usepackage{mwe}

\begin{document}

\title{
Focal Loss in 3D Object Detection
}

\author{
  Peng Yun$^{1}$ \
  Lei Tai$^{2}$ \
  Yuan Wang$^{2}$  \
  Chengju Liu$^{3}$ \
  Ming Liu$^{2}$ \
\thanks{This work was supported by the National Natural Science Foundation of China (Grant No. U1713211), and was partially supported by Shenzhen Science Technology and Innovation Commission (SZSTI) JCYJ20160428154842603, the Research Grant Council of Hong Kong SAR Government, China, under Project No. 11210017, No. 16212815 and No. 21202816 awarded to Prof. Ming Liu.
\textit{(Corresponding author: Peng Yun.)}}
\thanks{$^{1}$Peng Yun is with the Department of Computer Science and Engineering, The Hong Kong University of Science and Technology, Hong Kong (e-mail: {\tt\footnotesize pyun@ust.hk})
}
\thanks{$^{2}$Lei Tai, Yuan Wang and Ming Liu are with the Department of Electronic and Computer Engineering, The Hong Kong University of Science and Technology, Hong Kong (e-mail: {\tt\footnotesize \{ltai, ywangeq, eelium\}@ust.hk})}
\thanks{$^{3}$Chengju Liu is with College of Electrical and Information Engineering, Tongji University, China (e-mail: {\tt\footnotesize liuchengju@tongji.edu.cn})
}
}


\maketitle

\begin{strip}\centering
  \vspace{-4cm}
  \includegraphics[width=\textwidth]{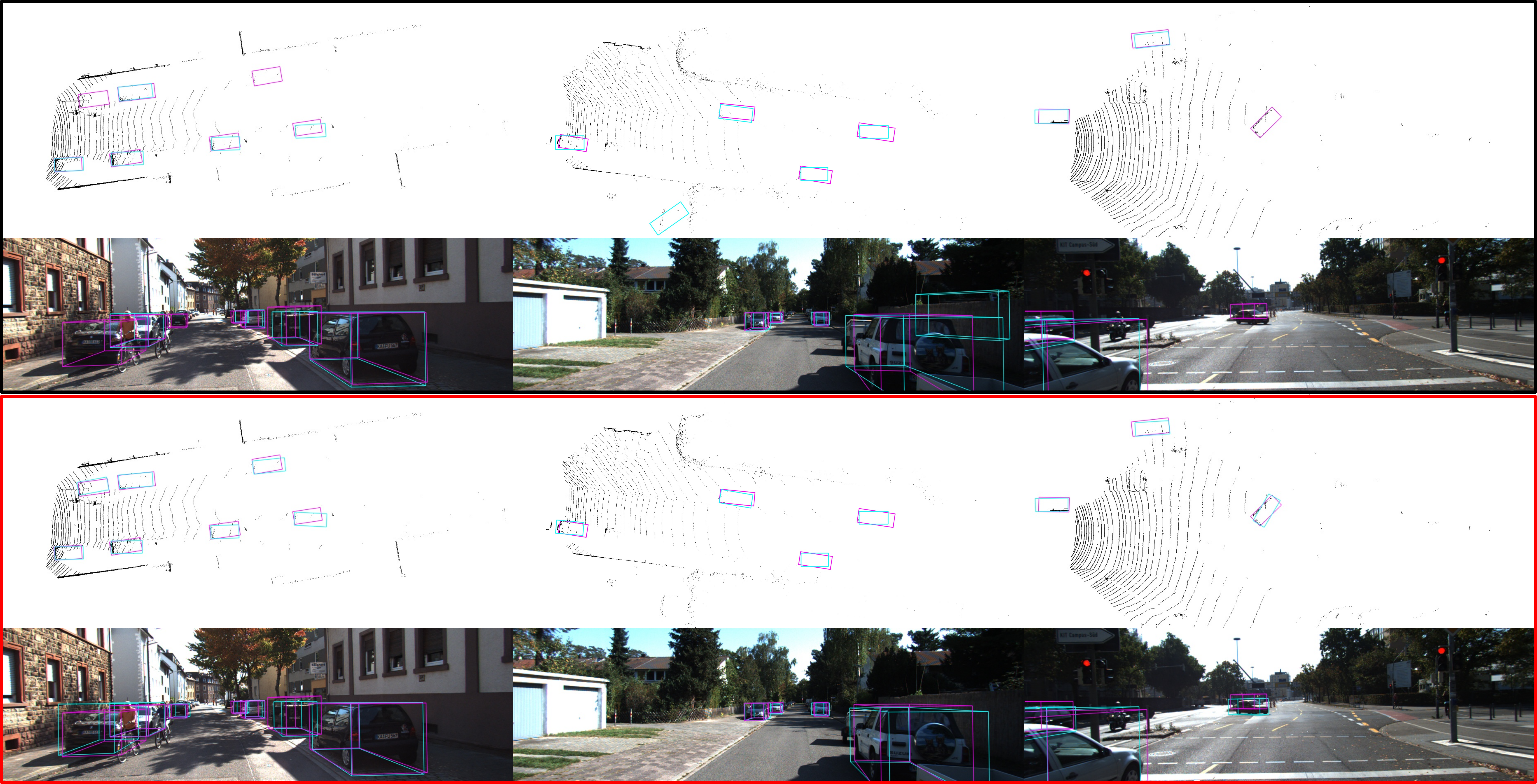}
  \captionof{figure}{
  Upper two rows show projected 3D object detection results from the detector trained with binary cross entropy. Lower two rows present related results from the detector trained with the focal loss.
  Purple and blue bounding boxes are the ground-truth and the estimated results respectively.
  \label{fig:feature-graphic}}
  \vspace{0.5cm}
\end{strip}


\begin{abstract}
3D object detection is still an open problem in autonomous driving scenes.
When recognizing and localizing key objects from sparse 3D inputs,
autonomous vehicles suffer from a larger continuous searching space and higher fore-background imbalance compared to image-based object detection.
In this paper, we aim to solve this fore-background imbalance in 3D object detection.
Inspired by the recent use of focal loss in image-based object detection,
we extend this hard-mining improvement of binary cross entropy to point-cloud-based object detection and conduct experiments to show its performance based on two different 3D detectors: 3D-FCN and VoxelNet.
The evaluation results show up to 11.2AP gains through the focal loss in a wide range of hyperparameters for 3D object detection.
\end{abstract}


\IEEEpeerreviewmaketitle

\input{secs/introduction}

\input{secs/related_work}
\input{secs/focal_loss}
\input{secs/3d_fcn}
\input{secs/voxelnet}
\input{secs/experiments}
\input{secs/conclusion}




\bibliographystyle{IEEEtran}
\bibliography{root}
\input{secs/appendix}
\end{document}

%% file: secs/introduction.tex
\vspace{-0.8cm}
\section{Introduction}
\label{sec:introduction}
\IEEEPARstart{O}{bject} detection in 3D is still challenging in robotics perception,
the applied scenes of which widely include urban and suburban roads, highways, bridges and indoor settings.
Robots recognize and localize key objects from data in the 3D form and predict their locations, sizes and orientations,
which provides both semantic and spatial information for high-level decision making.
The point cloud is one of the most commonly used 3D data forms, and can be gathered by range cameras, like LiDAR and RGB-D cameras.
Since the coordinate information of point clouds is not influenced by appearance changes,
point clouds are also robust in extreme weather and various seasons.
In addition, it is naturally scale-invariant.
The scale of an object is invariant anywhere in a point cloud,
while it always changes in an image due to foreshortening effects.
Moreover, the increasing perception distance and decreasing price of 3D LiDARs make them a promising direction for autonomous driving researchers \cite{wang2017characterization}.

Current image-based detectors benefit from translation invariance from convolution operations and can perform with human-comparable accuracy.
However, the successful image-based architectures cannot be directly applied in 3D space.
Point-cloud-based object detection consumes point clouds which are sparse point lists instead of dense arrays.
If drawing on the success of image-based detectors and conducting dense convolution operation to acquire translation invariance,
pre-processing must be implemented to convert the sparse point clouds into dense arrays.
Otherwise, special layers should be carefully designed to extract meaningful features from the sparse inputs.
Additionally, the fore-background imbalance is much more serious than in 2D scenarios,
since the new z-axis further enlarges the searching space and the extent of imbalance is different for each different z value.

Lin \textit{et al.}\cite{lin2018focal} proposed focal loss to tackle the fore-background imbalance in image-based object detection,
so that one-stage detectors could achieve state-of-the-art accuracy as two-stage detectors.
As a hard-mining improvement of binary cross entropy, it helps the network focus on hard classified objects, in case they are overwhelmed by a large number of easily classified objects.

Similar to image-based detection methods, point-cloud-based detection methods can also be classified into two-stage \cite{Qi_2018_CVPR,ku2017joint,chen2017multi} and one-stage detectors \cite{li20173d, Zhou_2018_CVPR}.
In this paper, inspired by \cite{lin2018focal}, we aim to solve the fore-background imbalance for 3D object detection through the focal loss.
We claim the following contributions:
\begin{itemize}
\item We extend focal loss to 3D object detection to solve the huge fore-background imbalance in one-stage detectors,
and conduct experiments on two different one-stage 3D object detectors, 3D-FCN \cite{li20173d} and VoxelNet \cite{Zhou_2018_CVPR}. 
The experiment results demonstrate up to 11.2AP gains from the focal loss in a wide range of hyperparameters.
\item To further understand focal loss in 3D object detection, we analyze its effect towards foreground and background estimations,
and validate that it plays a role similar to image-based detection.
We also find that the special architecture of VoxelNet can naturally handle the hard negatives well.
\item We plot the final posterior probability distributions of the two detectors and demonstrate that the focal loss with the increasing hyperparameter $\gamma$ decreases the estimation posterior probabilities.
\end{itemize}

%% file: secs/related_work.tex
\section{Related Work}
\label{sec:related_work}

\subsection{Two-Stage 3D Object Detection}
When extending two-stage image detectors to the 3D space, researchers encounter the following problems:
(1) the input is sparse and at low resolution;
(2) the original image-based methods are not guaranteed to have enough information to generate region proposals.
Ku \textit{et al.} \cite{ku2017joint} proposed AVOD which fused RGB images and point clouds.
It first proposes aligned 3D bounding boxes with a multimodal fusion region proposal network. Then,
the proposed bounding boxes are classified and regressed with fully connected layers.
Both the appearance and the 3D information are well-utilized to improve the accuracy and robustness of the proposed model in extreme scenes.
Their hand-crafted features can be further improved to learn representations directly from raw LiDAR inputs to alleviate information loss.

Qi \textit{et al.} \cite{Qi_2018_CVPR} proposed F-PointNet and leveraged both 2D object detectors and 3D deep learning for object localization.
They extracted the 3D bounding frustum of an object with a 2D object detector.
Then 3D instance segmentation and 3D bounding box regression were applied with two variants of PointNet \cite{qi2017pointnet}.
F-PointNet achieves state-of-the-art accuracy on the KITTI 3D object detection challenge \cite{Geiger2012CVPR}, and also performs at real-time speed for 3D object detection.
Their image detector needs to be carefully designed with a high recall rate, since the accuracy upper bound is determined by the first stage.

\subsection{One-Stage 3D Object Detection}
Li \cite{li20173d} extended a 2D fully convolutional network to 3D.
The voxelized point clouds are processed by an encoder-decoder network.
The 3D fully convolutional network (3D-FCN) finally proposes a probability and a regression map for the whole detection region.
It thoroughly consists of 3D dense convolutions with high computation and memory costs,
so that the network depth is limited and hard to extract high-level features.
Unlike 3D-FCN and AVOD, both of which adopt hand-crafted features to represent the point clouds,
Zhou \textit{et al.} \cite{Zhou_2018_CVPR} designed an end-to-end network to implement point-cloud-based 3D object detection with learning representations called VoxelNet.
Compared to 3D-FCN \cite{li20173d}, the computation cost is mitigated by the Voxel Feature Encoding Layers (VFELayers) and 2D convolution.

In this paper, we adopt 3D-FCN \cite{li20173d} and VoxelNet \cite{Zhou_2018_CVPR} as two different types of one-stage 3D detectors.
As shown in Table \ref{tab:comp}, 3D-FCN consumes dense grids and consists of only 3D dense convolution layers,
where the 2D FCN architecture \cite{long2015fully} is extended to 3D for dense feature extraction.
In contrast, VoxelNet consumes sparse point lists and is a heterogeneous network,
which firstly extracts sparse features with its novel VFELayers and then conducts 3D and 2D convolution sequentially.
\begin{table}[]
  \begin{center}
  \caption[caption]{Image-Based and Point-Cloud-Based Object Detection}
  \label{tab:comp}
  \renewcommand\arraystretch{1.1}
  \renewcommand\tabcolsep{3pt}
  \begin{tabular}{@{}cccc@{}}
  \toprule
  & \begin{tabular}[c]{@{}c@{}}Image-Based \\ Object Detection\end{tabular} & \multicolumn{2}{c}{\begin{tabular}[c]{@{}c@{}}Point-Cloud-Based \\ Object Detection\end{tabular}}                       \\ \midrule
  Method & - & 3D-FCN \cite{li20173d} & VoxelNet \cite{Zhou_2018_CVPR} \\
  Dimension & 2D                                                                               & 3D                                                           & 3D                                                                \\
  Input     & Dense Grid                                                                       & Dense Grid                                                   & 
  Sparse Point List
  \\
  Network &  Dense Conv & Dense Conv & Heterogeneous \\
  Pipeline & One/Two-Stage & One-Stage & One-Stage                                                       \\ \bottomrule
  \end{tabular}
  \end{center}
  \vskip -0.5cm
\end{table}
\subsection{Imbalance between Foreground and Background}


Image-based object detectors can be classified into two-stage and one-stage detectors.
For two-stage detectors, like R-CNN \cite{girshick2014rich},
the first stage generates a sparse set of candidate object locations
and the second stage classifies each candidate location as one of the foreground classes or as the background using a convolutional neural network.
The two-stage detectors \cite{he2017mask, lin2017feature} achieve state-of-the-art accuracy on the COCO benchmark.
On the other hand, one-stage detectors,
like YOLO \cite{redmon2016you} and SSD \cite{liu2016ssd}, aim to simplify the pipeline.
They improve the training speed of deep models and also demonstrate promising results in terms of accuracy.

Lin \textit{et al.} \cite{lin2018focal} explored both one-stage and two-stage detectors in image-based object detection,
and claimed that the hurdle that obstructs the one-stage detectors from better accuracy is the extreme fore-background class imbalance encountered during training of dense detectors.
They reshaped the standard cross entropy loss and proposed the focal loss such that the losses assigned to well-classified examples were down-weighted.
This can be seen as a hard-mining improvement of binary cross entropy to help networks focus on hard classified objects in case they are overwhelmed by a large number of easily classified objects.

We extend focal loss to 3D object detection to tackle the fore-background imbalance problem.
Different from image-based detection, point-cloud-based object detection is a more challenging perception problem in 3D space with sparse sensor data
and suffers from more serious fore-background imbalance.
To thoroughly evaluate the performance of the focal loss in this harder task, we conduct experiments based on two different types of one-stage 3D detectors: 3D-FCN and VoxelNet.
We analyze the focal loss effect on these two 3D detectors following a similar method to that in \cite{lin2018focal},
and further discuss the decreasing posterior probability effect of the focal loss.

%% file: secs/focal_loss.tex
\section{Focal Loss}
\label{sec:focal_loss}

In this section, we first declare notations and revisit the focal loss \cite{lin2018focal},
and then further analyze the fore-background imbalance in 3D object detection.

\subsection{Preliminaries}
We define $y \in \{\pm 1\}$ as the ground-truth class,
and $p$ as the estimated probability for the class with label $y=1$.
For notational convenience, we define the posterior probability $p_t$ as
\begin{equation}
    p_t =
    \begin{cases}
        p       & if\ y=1\\
        1 - p   & if\ y=-1,
    \end{cases}
\end{equation}
where p is calculated with $p = sigmoid(x)$. 
The binary cross entropy (BCE) loss and its deviation can be formulated as
\begin{align}
    \epsilon_{BCE} \left(p_t\right) &= -\log\left(p_t\right)
    \label{eq:bce_loss}\\
    \frac{d\epsilon_{BCE}\left(p_t\right) }{dx}  &= y\left(p_t -1\right).
\end{align}

As claimed in \cite{lin2018focal},
when the network is trained with BCE loss, its gradient will be dominated by vast easy classified negative samples
if a huge fore-background imbalance exists.
Focal loss can be considered as a dynamically scaled cross entropy loss, which is defined as
\begin{align}
    &\epsilon_{FL}\left(p_t\right) = -\left(1-p_t\right)^\gamma\log\left(p_t\right)
    \label{eq:fl}\\
    \frac{d\epsilon_{FL}\left(p_t\right) }{dx} &= y\left(1-p_t\right)^\gamma\left(\gamma p_t log\left(p_t\right) + p_t -1\right).
\end{align}

The contribution from the well classified samples ($p_t \gg 0.5$) to the loss is down-weighted.
The hyperparameter $\gamma$ of the focal loss can be used to tune the weight of different samples.
As $\gamma$ increases, fewer easily classified samples contribute to the training loss.
Obviously, when $\gamma$ reaches $0$, the focal loss degrades to become same as the BCE loss.
In the following sections, all the cases with $\gamma=0$ represent BCE loss cases.

Researchers have previously either introduced hyperparameters to balance the losses calculated from positive and negative anchors,
or normalized positive and negative losses by the frequency of corresponding anchors.
However, one essential problem that these two previous methods cannot handle is the gradient salience of hard negative samples.
The gradients of hard negative anchors ($p_t<0.5$) are overwhelmed by a large number of easy negative anchors ($p_t\gg0.5$).
Due to the dynamic scaling with the posterior probability $p_t$, a weighted focal loss can be used to handle both the fore-background imbalance and the gradient salience of hard negative samples with the following form,
\begin{align}
    &\epsilon_{FL}\left(p_t\right) = -\lambda\left(1-p_t\right)^\gamma\log\left(p_t\right),
    \label{eq:fl}
\end{align}
where $\lambda$ is induced to weight different classes.
In the following sections, we adopt hyperparameters $\alpha$ and $\beta$ to weight positive and negative focal loss respectively.

\subsection{Fore-background Imbalance in 3D Object Detection}
The methods for 3D object detection can be classified as one-stage \cite{li20173d, Zhou_2018_CVPR} and two-stage \cite{Qi_2018_CVPR,ku2017joint,chen2017multi} detectors.
The two-stage detectors first adopt an algorithm with a high recall rate to propose regions that possibly contain objects
and adopt a convolution network to classify classes and regress bounding boxes.
The one-stage detectors are end-to-end networks that learn representations and implement classification and regression in all anchors.

In one-stage methods, anchors are proposed at each location,
and thus a huge fore-background imbalance exists.
For instance, there are 50k bounding boxes proposed in each frame for 3D-FCN and 70k for VoxelNet,
but less than 30 anchors among them contain positive objects (e.g. car, pedestrian, cyclist).
Compared to image detectors, the extra estimation in z-axis further increases the fore-background imbalance.
Additionally, positive samples always locate on the position with small z values in some specific scenes.
For instance, cars and pedestrians are always on the road in autonomous driving scenes.
In such situations, the distribution of fore-background imbalance is different along the z-axis:
the extent of imbalance increases with higher z values.

The one-stage methods for 3D detectors are different from the 2D detectors because of their larger searching space, 
sparse input and different types of network architecture.
Therefore, we select two different networks, 3D-FCN and VoxelNet, to conduct experiments to evaluate the performance of focal loss in 3D object detection.
The features of these two 3D detectors are discussed in the following two sections, and the experimental details and results are shown in Section \ref{sec:experiment}.


%% file: secs/3d_fcn.tex
\section{3D-FCN Features}
\label{sec:3d_fcn}
In this section, we discuss the dense convolution network architecture of 3D-FCN and introduce our enhanced loss function for 3D-FCN.
The details of 3D-FCN can be found in \cite{li20173d}.
Please refer to APPENDIX for our implementation of 3D-FCN.

\subsection{Dense Convolution Network Architecture}
\begin{figure}[]
    \centering
    \includegraphics[width=\linewidth]{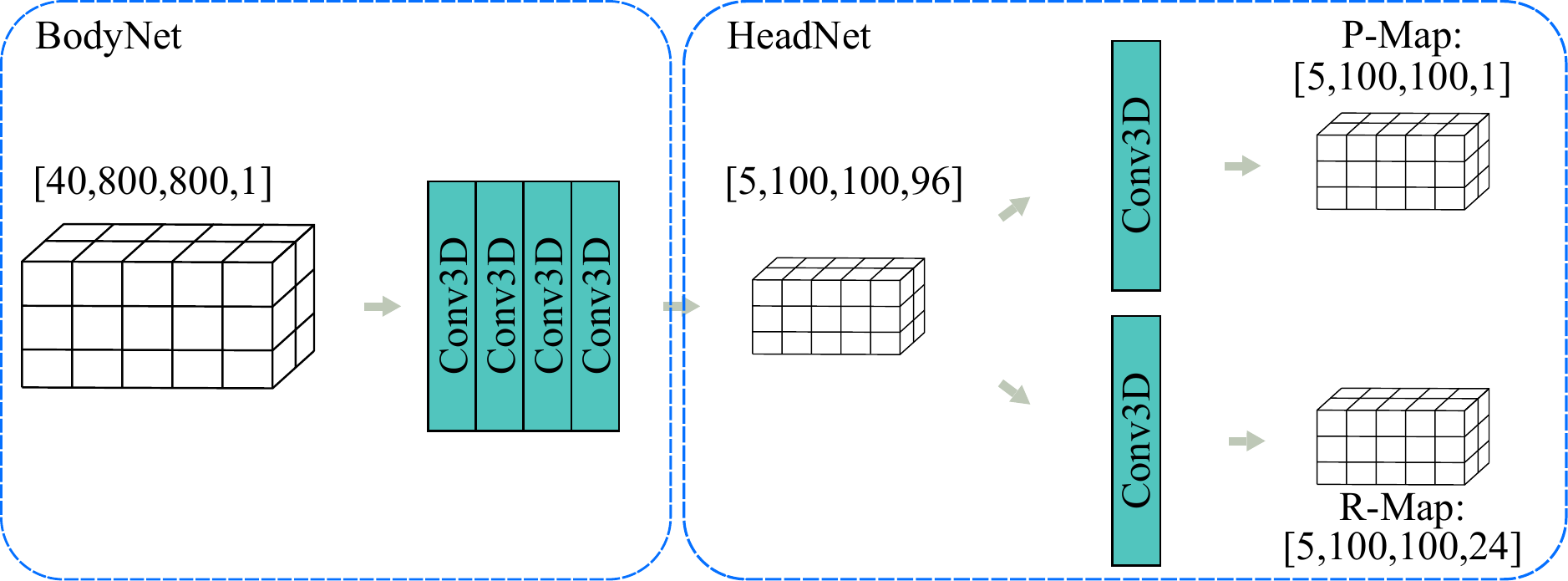}
    \caption{The dense convolution network architecture of 3D-FCN \cite{li20173d}.
    The whole network consists of only 3D convolution layers.
    All intermediate tensors in the hidden space are dense 3D grids (which are represented by a tensor with dimensions as [height, width, length, feature]).
    }
    \vskip -0.6cm
    \label{fig:3dfcn-arc}
 \end{figure}
 3D-FCN \cite{li20173d} draws on experience from image-based recognition tasks,
 and extends the 2D convolution layer to 3D space to acquire translation invariance.
 The input point cloud is firstly voxelized into a 3D dense grid.
 In each voxel of the 3D dense grid, the values $\{0,1\}$ are used to present whether there is any point observed.
 The network architecture of 3D-FCN is shown in Figure \ref{fig:3dfcn-arc}.
 The voxelized point cloud is convolved by four Conv3D blocks sequentially.
 The output features are then processed by two Conv3D blocks individually to generate a probability map and a regression map (P-Map and R-Map).
 Different from image-based object detection, the probability map and regression map are all in 3D dense grids,
 so that the searching space is exponentially increased.

\subsection{Enhanced Loss Function}
The original loss function for 3D-FCN \cite{li20173d} is shown in the left of Equation \ref{eq:3dfcn_loss_function1} to \ref{eq:3dfcn_loss_function5},
where $\epsilon_P$ and $\epsilon_R$ represent the classification loss and regression loss,
as well as $\epsilon_{cls}$ and $\epsilon_{reg}$ are the loss functions used for classification and regression respectively.
In regression loss $\epsilon_R$,  $\bf u_i$ and $\bf u_i^*$ are the regression output and ground truth for positive anchors.
In classification loss $\epsilon_P$, $p_i^{pos}$ and $p_i^{neg}$ represent the posterior probability of positive and negative estimation.

\begin{align}
    &\epsilon = \epsilon_P + \epsilon_R                                  \hspace{1.4cm}  \rightarrow \hspace{0.05cm} \epsilon = \epsilon_P + \epsilon_R  \label{eq:3dfcn_loss_function1}\\
    &\epsilon_P = \eta (\epsilon_{P}^{pos} + \epsilon_{P}^{neg})         \hspace{0.25cm} \rightarrow \hspace{0.05cm} \epsilon_P = \eta (\epsilon_{P}^{pos} + \epsilon_{P}^{neg}) \label{eq:3dfcn_loss_function2}\\    
    &\epsilon_R = \sum_{i} \epsilon_{reg}(\mathbf{u_i}, \mathbf{u_i^*})  \hspace{0.35cm} \rightarrow \hspace{0.05cm} \epsilon_R = \frac{1}{N_{pos}} \sum_{i} \epsilon_{reg}(\mathbf{u_i}, \mathbf{u_i^*}) \label{eq:3dfcn_loss_function3}\\
    &\epsilon_{P}^{pos} =  \sum_{i} \epsilon_{cls}({p_i^{pos}}, 1)                       \rightarrow \hspace{0.05cm} \epsilon_{P}^{pos} = \alpha \frac{1}{N_{pos}} \sum_{i} \epsilon_{cls}({p_i^{pos}}, 1) \label{eq:3dfcn_loss_function4}\\
    &\epsilon_{P}^{neg} =  \sum_{i} \epsilon_{cls}({p_i^{neg}}, 0)                       \rightarrow \hspace{0.05cm} \epsilon_{P}^{neg} = \beta  \frac{1}{N_{neg}} \sum_{i} \epsilon_{cls}({p_i^{neg}}, 0) \label{eq:3dfcn_loss_function5}
\end{align}

In the original form, a large imbalance exists between $\epsilon_{P}^{pos}$ and $\epsilon_{P}^{neg}$,
which represent classification loss of positive and negative samples respectively.
Therefore, we adopt the loss function used in VoxelNet \cite{Zhou_2018_CVPR}, which normalizes sub-loss with corresponding frequency as well as balances $\epsilon_{P}^{pos}$ and $\epsilon_{P}^{neg}$ with two more hyperparameters $\alpha$ and $\beta$.
The adopted loss function is shown in the right of Equation \ref{eq:3dfcn_loss_function1} to \ref{eq:3dfcn_loss_function5}.

In Section \ref{sec:experiment}, we use the loss function in the right part of Equation \ref{eq:3dfcn_loss_function1} to \ref{eq:3dfcn_loss_function5} to demonstrate the focal loss improvement compared with BCE Loss,
where $\epsilon_{reg}$ denotes the square loss and $\epsilon_{cls}$ denotes the focal loss.
We also show the enhanced loss function form improvement compared with the original loss function \cite{li20173d} in the APPENDIX,
where $\epsilon_{reg}$ denotes the square loss and $\epsilon_{cls}$ denotes the BCE loss. 

%% file: secs/voxelnet.tex
\section{VoxelNet Features}
\label{sec:voxelnet}
In this section, we discuss the heterogeneous network architecture of VoxelNet,
and its bird's-eye-view estimation.
The details of VoxelNet can be found in \cite{Zhou_2018_CVPR}.
Please refer to APPENDIX for our implementation of VoxelNet.

\subsection{Heterogeneous Network Architecture}
\begin{figure*}[t]
    \begin{center}
    \includegraphics[width=\linewidth]{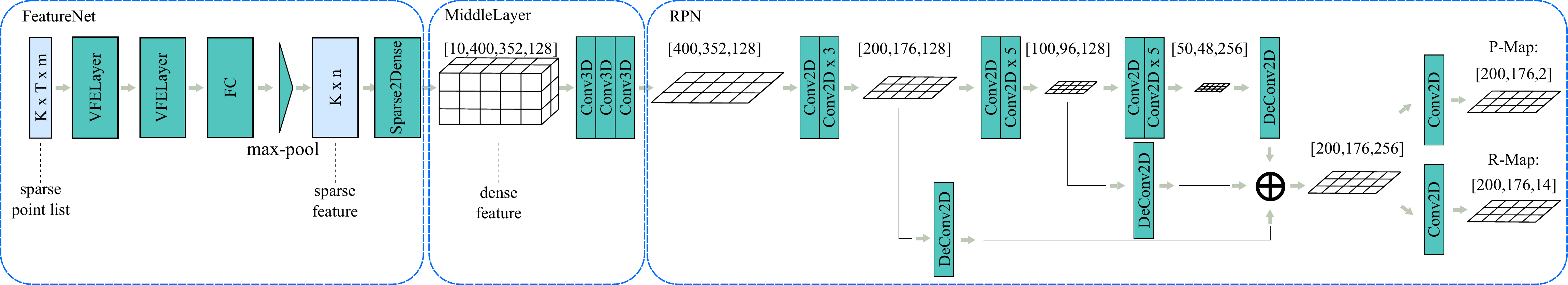}
    \caption{VoxelNet heterogeneous architecture \cite{Zhou_2018_CVPR}.
    It consists of three main parts: FeatureNet (point-wise and voxel-wise feature transformation), MiddleLayer (3D dense convolution) and RPN (2D dense convolution).
    The probability and regression maps are in bird's-eye-view form.
    }
    \vskip -0.7cm
    \label{fig:voxelnet_architecture}
    \end{center}
\end{figure*}

The heterogeneous architecture overview of VoxelNet is shown in Figure \ref{fig:voxelnet_architecture}.
It consists of three main parts: FeatureNet, MiddleLayer and RPN.

\textit{FeatureNet} extracts features directly from sparse point lists.
It adopts Voxel Feature Encoding Layers (VFELayers) \cite{Zhou_2018_CVPR} to extract both point-wise and voxel-wise features directly from points,
where fully connected layers are used to extract point-wise features and a symmetric function is used to aggregate local features from all points within a local voxel.
Compared to sub-optimally deriving hand-crafted features from voxels,
VFELayers can learn representations minimizing the loss function.
The derived voxel-wise representations from VFELayers are sparse,
which saves memory and time in the computation.
In contrast, if a point cloud of KITTI dataset is partitioned into a $[10,400,352]$ dense grid for vehicle detection,
only around 5300 voxels (about 0.3\%) are non-empty.
However, the sparse representation is currently unfriendly to convolutional operations.
In order to implement convolution, VoxelNet compromises on efficiency and converts the sparse representation
to a dense representation at the end of FeatureNet. Each sparse voxel-wise representation is copied to its specific entry in the dense grid.

\textit{MiddleLayer} consumes the 3D dense grid and converts it to a 2D bird's-eye-view form,
so that further processing can be done in 2D space.
The role of MiddleLayer is to learn features from all voxels in the same bird's-eye-view location.
Therefore, the 3D convolutional kernel is of size $[d, 1, 1]$,
if we denote the dense grid in the order of $z, x, y$.
The 3D kernel of size $[d, 1, 1]$ helps aggregate voxel-wise features within a progressively expanding receptive field along the z-axis
and keeps the shape in the $x,y$ dimension.

\textit{RPN} predicts the probability and regression map from the 2D bird's-eye-view feature map.
Since the increased invariance and large receptive fields of top-level nodes will yield smooth responses and cause inaccurate localization,
it does not utilize max-pooling but adopts skip-layers \cite{long2015fully} to combine high-level semantic features and low-level spatial features.

\subsection{Estimation in Bird's-Eye-View Form}
The final probability and regression estimation maps are all in bird's-eye view form,
which is similar to the final estimation of image-based detection methods.
This saves both memory and time of the calculation compared to 3D maps, but only one object per location can be estimated in the bird's-eye view.
This is acceptable in autonomous driving scenes but will meet problems in indoor scenes, where objects can be stacked up (e.g., a mug on a stack of books).

MiddleLayer saves calculation for further processing by aggregating the 3D dense grid into a 2D bird's-eye-view feature map.
Otherwise, thoroughly 3D dense convolution in such a deep network (22 convolution layers) would bring exponentially more parameters and calculation.
We note that MiddleLayer is still a bottleneck of the whole network as shown in Table \ref{tab:network_details_voxelnet} because of its
3D dense convolution operations.
The efficient sparse convolutional implementation is still an open problem and deserves effort to solve.

\subsection{Loss Function}
We adopt the loss function form from the original VoxelNet \cite{Zhou_2018_CVPR}, which is the same as the right half part from Equation \ref{eq:3dfcn_loss_function1} to \ref{eq:3dfcn_loss_function5}.
In Section \ref{sec:experiment}, we use SmoothL1Norm \cite{ren2017faster} for $\epsilon_{reg}$ as the original paper \cite{Zhou_2018_CVPR} and use the focal loss for $\epsilon_{cls}$.

%% file: secs/experiments.tex
\section{Experiments}
\label{sec:experiment}
\begin{figure*}[h]
    \centering
    \includegraphics[width=0.85\textwidth]{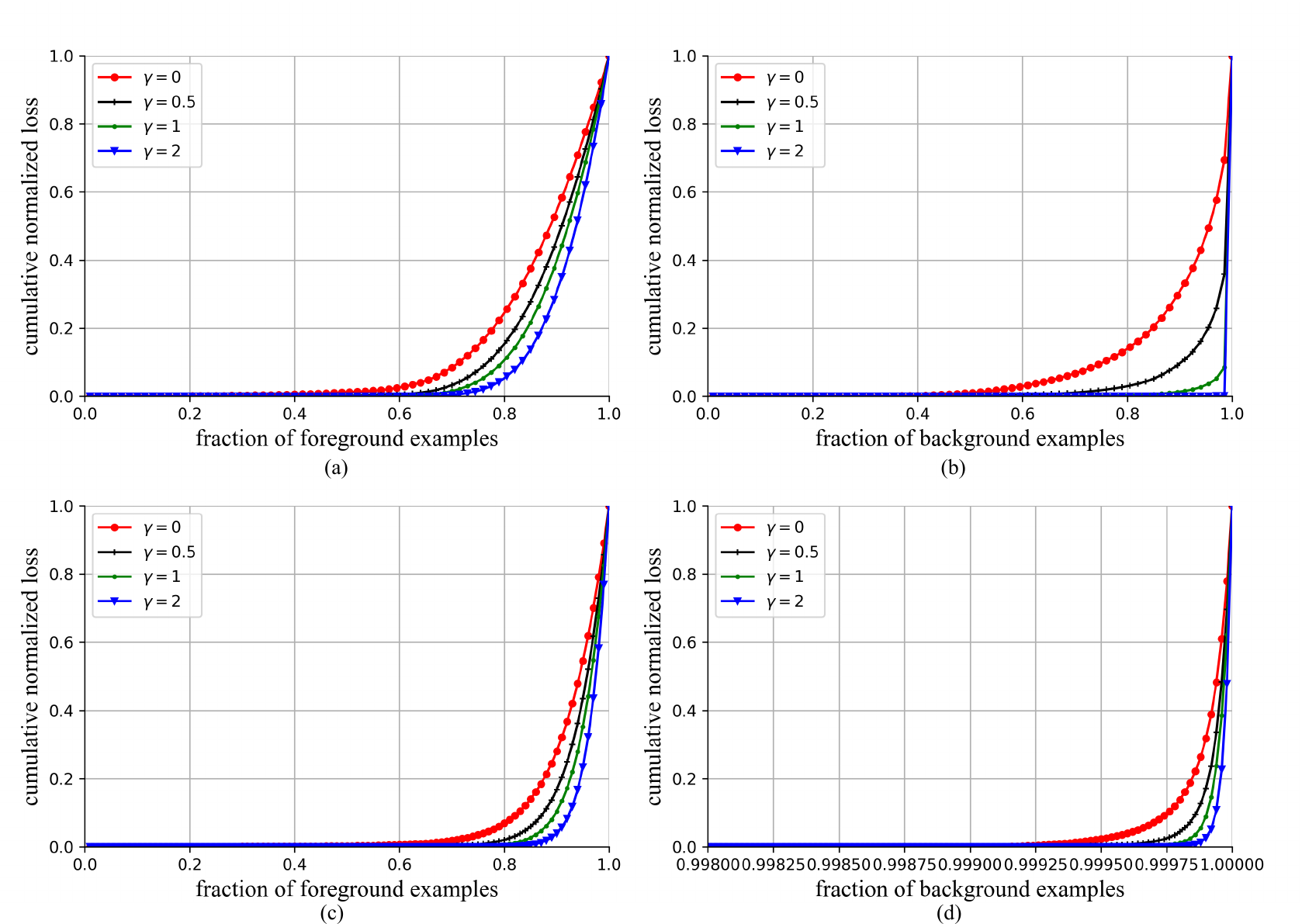}
    \caption{Cumulative distributions of 3D-FCN and VoxelNet for different values of $\gamma$.
    In 3D-FCN (a, b), as $\gamma$ increases, loss of both foreground and background samples concentrate on the harder partitions. The effect on the background is stronger.
    In VoxelNet (c, d), the effect of the focal loss increases as $\gamma$ increases, but the effect on the foreground is stronger than on the background.
    Note that the VoxelNet background cumulative distribution (d) is in the range of $[0.998, 1]$.
    }
    \label{fig:cdf}
 \end{figure*}

\begin{table*}[thbp]
    \begin{minipage}{.5\linewidth}
    \centering
    \caption{Evaluation Results on KITTI Validation Dataset \protect\\For Last Models of 3D-FCN}
    \label{tab:3d_fcn_comp}
    \renewcommand\arraystretch{1.1}
    \renewcommand\tabcolsep{8pt}
    \begin{tabular}{@{}ccccccc@{}}
        \toprule
        \multirow{2}{*}{$\gamma$}       & \multicolumn{3}{c}{Bird's Eye View AP (\%)}      & \multicolumn{3}{c}{3D Detection AP (\%)}        \\
                                        & Easy           & Mod            & Hard           & Easy           & Mod            & Hard  \\ \midrule
        0                               & 32.11          & 31.67          & 27.78          & 24.22          & 21.96          & 18.63          \\
        0.1                             & \textbf{37.53} & \textbf{35.15} & \textbf{30.61} & \textbf{28.24} & \textbf{24.73} & \textbf{24.80} \\
        0.2                             & \textbf{38.10} & \textbf{35.32} & \textbf{30.65} & \textbf{27.75} & \textbf{23.88} & \textbf{20.36} \\
        0.5                             & \textbf{33.59} & \textbf{32.61} & \textbf{28.59} & \textbf{24.76} & \textbf{22.34} & \textbf{19.04} \\
        1                               & \textbf{42.91} & \textbf{38.21} & \textbf{32.96} & \textbf{32.26} & \textbf{26.70} & \textbf{22.58} \\
        2                               & \textbf{43.32} & \textbf{38.45} & \textbf{33.09} & \textbf{32.91} & \textbf{27.23} & \textbf{22.81} \\
        5                               & 25.18          & 24.38          & 20.62          & 18.77          & 16.47          & 17.27          \\ \bottomrule
        \end{tabular}
    \end{minipage}\begin{minipage}{.5\linewidth}
    \centering
    \caption{Evaluation Results on KITTI Validation Dataset \protect\\For Last Models of VoxelNet}
    \label{tab:voxelnet_comp}
    \renewcommand\arraystretch{1.1}
    \renewcommand\tabcolsep{8pt}
    \begin{tabular}{@{}ccccccc@{}}
        \toprule
        \multirow{2}{*}{$\gamma$}       & \multicolumn{3}{c}{Bird's Eye View AP (\%)}      & \multicolumn{3}{c}{3D Detection AP (\%)}        \\
                                        & Easy           & Mod            & Hard           & Easy           & Mod            & Hard  \\ \midrule
        0                               & 85.26          & 61.35          & 60.97          & 70.54          & 55.11          & 48.79          \\
        0.1                             & \textbf{85.93} & \textbf{69.65} & \textbf{68.83} & \textbf{72.67} & \textbf{56.31} & \textbf{56.11} \\
        0.2                             & 82.55          & 60.42          & 60.23          & \textbf{72.66} & \textbf{56.67} & \textbf{50.41} \\
        0.5                             & \textbf{86.80} & \textbf{69.40} & \textbf{61.79} & \textbf{75.86} & \textbf{58.28} & \textbf{57.92} \\
        1                               & \textbf{87.28} & \textbf{70.46} & \textbf{61.93} & \textbf{74.16} & \textbf{57.01} & \textbf{56.20} \\
        2                               & 84.48          & \textbf{68.76} & \textbf{61.04} & \textbf{70.82} & \textbf{55.25} & \textbf{54.67} \\
        5                               & 80.48          & \textbf{62.56} & 53.76          & \textbf{75.04} & 50.85          & \textbf{50.53} \\ \bottomrule
        \end{tabular}
    \end{minipage}
    \vskip -0.5cm
\end{table*}

In this section, we intend to answer two questions:
1) Can focal loss help improve accuracy in 3D object detection task?
2) Does focal loss have an equal effect in 3D object detection to its effect in image-based detection?
To answer the former question, we conduct experiments to compare the performance of 3D-FCN and VoxelNet trained with BCE loss and focal loss on the challenging KITTI benchmark \cite{Geiger2012CVPR}.
To answer the second question, we analyze the cumulative distribution curve of 3D-FCN and VoxelNet following a similar method to that in \cite{lin2017feature}.
The code and weights for our experiments are available at \url{https://sites.google.com/view/fl3d}.

\subsection{BCE Loss vs. Focal Loss}
The KITTI 3D object detection dataset \cite{Geiger2012CVPR} contains 3D annotations for cars, pedestrians and cyclists in urban driving scenarios.
The sensor setup mainly consists of a wide-angle camera and a Velodyne LiDAR (HDL-64E), both of which are well-calibrated.
The training dataset contains 7481 frames, including both raw sensor data and annotations.
The KITTI 3D detection dataset contains some bad annotations which are empty bounding boxes containing few points.
In order to avoid overfitting those bad annotations, we remove all bounding boxes containing few points (fewer than 10).
Following \cite{chen2017multi}, we split the dataset into training and validation sets, each containing around half of the entire set.

For simplicity, we conduct experiments only on the car class to show the focal loss improvement.
We do such implement because both 3D-FCN and VoxelNet are trained class-specifically and extending them to other classes is only tuning techniques.
Also, the focal loss in the form of Equation \ref{eq:fl} is agnostic to the class of objects.

We set $\alpha=1$, $\beta=5$, $\eta=10$ in 3D-FCN and $\alpha=1$, $\beta=10$, $\eta=0.5$ in VoxelNet
so that $\epsilon_{P}^{pos}$ and $\epsilon_{P}^{neg}$ as well as $\epsilon_{P}$ and $\epsilon_{R}$ will be of the same orders of magnitude.
As claimed in \cite{lin2018focal}, when training a network from scratch with the focal loss, it is unstable in the beginning.
Therefore,
we first train the network (both 3D-FCN and VoxelNet) for 30 epochs with the BCE loss and the learning rate $lr$,
and then for another 30 epochs with the focal loss and a discounted learning rate $0.1lr$.
The minimum overlap thresholds are $0.7, 0.5, 0.5$ for 2D evaluation on image/ground plane and 3D evaluation.
The network details of both 3D-FCN and VoxelNet are shown in Table \ref{tab:network_details_3dfcn} and Table \ref{tab:network_details_voxelnet} in APPENDIX.
Non-maximum suppression with the threshold $0.8$ is used at the end of 3D-FCN and VoxelNet for estimation refinement.

In order to control a single variable $\gamma$, we firstly make comparisons among last models,
which are trained with the same amount of steps.
Additionally, we also make comparisons among best models to make the conclusion more concrete.
The best models are selected according to the mean value among easy, moderate and hard 3D detection APs (3D detection mAP).

We compare the results of the last models in Table \ref{tab:3d_fcn_comp} and Table \ref{tab:voxelnet_comp},
where the rows with $\gamma = 0$ and $\gamma >0$ represent the results from the BCE loss
and the focal loss respectively.
Bolded numbers are the results in which focal loss cases outperforms the BCE loss case.
In general, VoxelNet outperforms 3D-FCN in accuracy,
since the input of VoxelNet has the original point clouds, but 3D-FCN suffers from information loss when voxelizing the point clouds into binary representations.
Additionally, VoxelNet benefits from its deeper network structure, which is able to extract more useful high-level features.
In 3D-FCN, the focal loss helps improve accuracy in all metrics in a wide range of hyperparameters ($0 < \gamma \leq 2.0$), providing gains from 0.3AP to 11.2AP.
In VoxelNet, the cases with $\gamma = 0.1, 0.5, 1$ show gains from the focal loss in all metrics, ranging from 0.6AP to 9.1AP.
Both gains and losses happen when $\gamma$ is $0.2$ or $2$.
However, gains (up to 9.1AP) are generally much greater than losses (at most 2.7AP).
The training processes include some randomness due to sample shuffling and the sophisticated gradient descent training scheme.
We further evaluate all intermediate weights and select the best models to make the comparison in Table \ref{tab:best_result}.
It shows that focal loss helps improve accuracy in all metrics with a proper $\gamma$.
The performance losses of $\gamma = 0.2$ in Table \ref{tab:voxelnet_comp} might be caused by training randomness and model degradation with redundant training.

From Table \ref{tab:3d_fcn_comp}, Table \ref{tab:voxelnet_comp} and Table \ref{tab:best_result},
it shows that the focal loss in 3D object detection provides better or comparable results than BCE loss.
Therefore, the focal loss works in 3D object detection and help improve accuracy in a wide range of $\gamma$ (normally $\gamma \leq 2$).

\subsection{Analysis of Focal Loss in 3D Detectors}
\begin{table}[]
    \begin{center}
    \caption{Evaluation Result on KITTI Validation Dataset \protect\\ For Best Models}
    \label{tab:best_result}
    \renewcommand\arraystretch{1.1}
    \renewcommand\tabcolsep{3pt}
    \begin{tabular}{@{}cccccccccc@{}}
        \toprule
        \multirow{2}{*}{Detector}          & \multirow{2}{*}{$\gamma$}          & \multirow{2}{*}{lr}          & \multirow{2}{*}{Step}          & \multicolumn{3}{c}{\begin{tabular}[c]{@{}c@{}}Bird's Eye View\\ AP(\%)\end{tabular}}          & \multicolumn{3}{c}{\begin{tabular}[c]{@{}c@{}}3D Detection\\ AP(\%)\end{tabular}} \\
                                           &                                    &                              &                                & Easy                          & Mod                           & Hard                          & Easy                         & Mod                          & Hard                \\ \midrule
        3D-FCN                             & 0                                  & 1e-2                         & 126k                           & 51.33                         & 45.82                         & 40.24                         & 40.01                        & 33.12                        & 28.94                        \\
        3D-FCN                             & 2                                  & 1e-2                         & 137k                           & \textbf{53.19}                & \textbf{48.03}                & \textbf{41.96}                & \textbf{46.05}               & \textbf{35.93}               & \textbf{31.01}               \\
        VoxelNet                           & 0                                  & 1e-4                         & 134k                           & 85.76                         & 68.52                         & 61.00                         & 75.42                        & 58.09                        & 57.61                        \\
        VoxelNet                           & 0.2                                & 1e-4                         & 215k                           & \textbf{86.89}                & \textbf{69.33}                & \textbf{61.63}                & \textbf{80.08}               & \textbf{58.39}               & 57.60                        \\ \bottomrule
        \end{tabular}
    \end{center}
    \vskip -0.8cm
    \bigskip
    \bigskip
    Note that all cases in Table \ref{tab:best_result} are the evaluation results of the best models selected among all intermediate weights.
    Thus the accuracy improvement is from the focal loss instead of longer training steps.
    \vskip -0.5cm
\end{table}

We analyze the empirical cumulative distributions of the loss from the converged 3D-FCN and VoxelNet models as in \cite{lin2018focal}.
\begin{figure*}[!th]
    \centering
    \includegraphics[width=\textwidth]{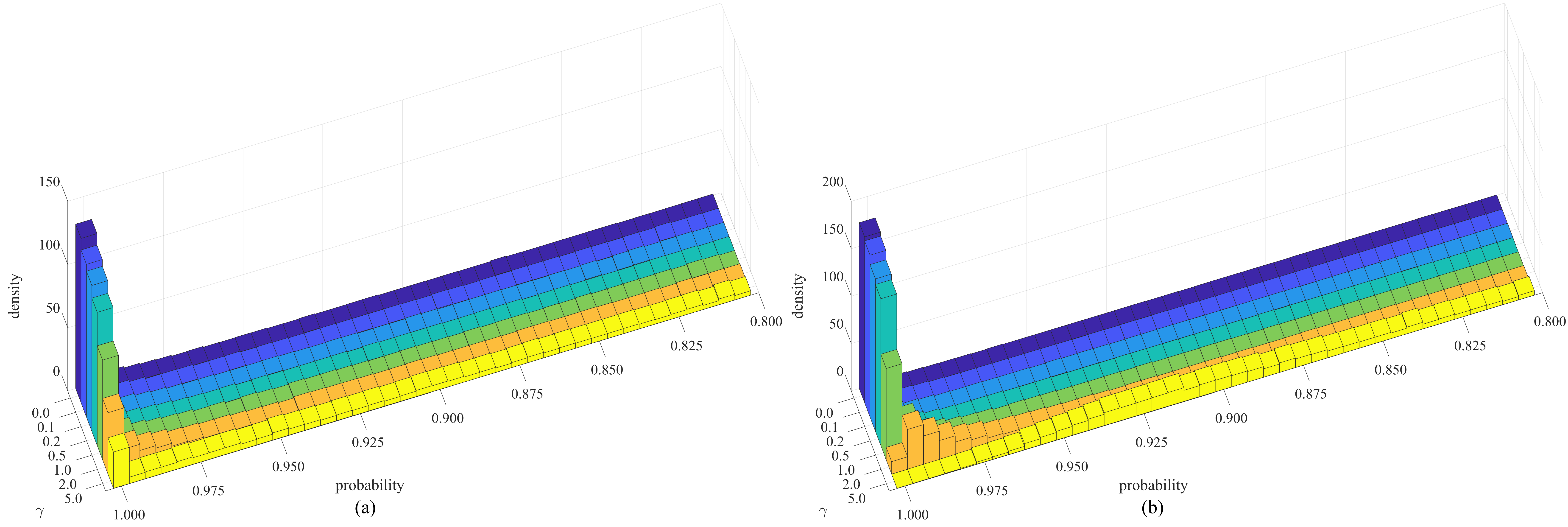}
    \caption{Posterior probability histogram of 3D-FCN (a) and VoxelNet (b).
    As $\gamma$ increases, the peak decreases and moves towards lower values in both 3D-FCN and VoxelNet.
    }
    \label{fig:confident}
    \vskip -0.4cm
 \end{figure*}
We apply the two converged models trained with the focal loss (row 2 and row 4 in Table \ref{tab:best_result}) on the validation dataset and sample the predicted probability for $10^7$ negative windows and $10^5$ positive windows.
Then, we calculate the focal loss with these probability data.
The calculated focal loss is normalized such that it sums to one and is sorted from low to high.
We plot the cumulative distributions for 3D-FCN and VoxelNet for different $\gamma$ in Figure \ref{fig:cdf}.


In 3D-FCN, approximately 15\% of the hardest positive samples account for roughly half of the positive loss.
As $\gamma$ increases, more of the loss gets concentrated in the top 15\% of examples.
However, compared to the effect of the focal loss on negative samples, its effect on the positive samples is minor.
For $\gamma=0$, the positive and negative CDFs are quite similar.
As $\gamma$ increases, more weight becomes concentrated on the hard negative examples.
With $\gamma=2$ (the best result for 3D-FCN), the vast majority of the loss comes from a small fraction of samples.
As claimed in \cite{lin2018focal}, the focal loss can effectively discount the effect of easy negatives,
so that the network focuses on learning the hard negative examples.

In VoxelNet, the condition is different.
From c and d in Figure \ref{fig:cdf}, we can see that the effect of the focal loss increases in both the positive and negative samples as $\gamma$ increases.
However, the cumulative distribution functions for the negative samples are quite similar among different values of $\gamma$,
even though we adjust the x-axis to $[0.998, 1]$.
This shows that VoxelNet trained with the BCE loss is already able to handle negative hard samples.
Compared with the results on the negative samples, the effects of focal loss on the positive samples are stronger.
Therefore, the accuracy gains of the focal loss in VoxelNet are mainly from the positive hard samples.

From the analysis of cumulative distributions,
we believe that the focal loss in 3D object detection helps networks alleviate hard sample gradient salience in the training process.

\subsection{Focal Loss Decreases the Posterior Probabilities}
When undertaking the experiments, we found networks trained with the focal loss should be set with a lower threshold for non-maximum suppression.
This inspires us to explore the influence of the focal loss on the output posterior probabilities.
We take the models in Table \ref{tab:3d_fcn_comp} and Table \ref{tab:voxelnet_comp}, and evaluate them on the validation set.
We record all the evaluation results and plot the probability histogram for positive bounding boxes.
The results are shown in Figure \ref{fig:confident}.
As $\gamma$ increases, the peak decreases and moves towards the lower values.
This demonstrates that networks trained with the focal loss output positive estimation with lower posterior probabilities.
A probable explanation is that objects with high posterior probabilities are easily classified, and the loss they contribute is down-weighted in the training process due to the focal loss.
In other words, they will be relatively ignored in the training process if they are estimated with high posterior probabilities, so that their posterior probabilities cannot be further improved.
However, they can also be accurately classified if we decrease the non-maximum suppression threshold in the final output step.

%% file: secs/conclusion.tex
\section{Conclusion}
\label{sec:conclusion}
In this paper, we extended the focal loss of image detectors to 3D object detection
to solve the fore-background imbalance.
We conducted experiments on two different types of 3D object detectors
to demonstrate the performance of the focal loss in point-cloud-based object detection.
The experimental results show that the focal loss helps improve accuracy in 3D object detection,
and it protects the network from fore-background imbalance and alleviates hard sample gradient salience both for positive and negative anchors in the training process.
The posterior probability histograms show that the networks trained with the focal loss outputs positive estimation with lower posterior probabilities.

%% file: secs/appendix.tex
\section*{APPENDIX}
\label{sec:appendix}
\subsection{Improvement of Enhanced Loss Function for 3D-FCN}

We demonstrate the improvement of adopting the loss function from VoxelNet \cite{Zhou_2018_CVPR} (normalization, new hyperparameters, BCE Loss)
provides over its original loss function \cite{li20173d} for 3D-FCN.
We set $\alpha=1$, $\beta=5$, $\eta=10$ in the enhanced 3D-FCN so that $\epsilon_{P}^{pos}$ and $\epsilon_{P}^{neg}$ as well as $\epsilon_{P}$ and $\epsilon_{R}$ can be of the same orders of magnitude.
We set $\eta=0.1$ in the original 3D-FCN so that $\epsilon_{P}$ and $\epsilon_{R}$ can be of the same orders of magnitude.
$\gamma$ is set as $0$ for using BCE loss.
We train these two cases from scratch with 30 epochs.
The threshold for non-maximum suppression is set as 0.994.
The reason why $\eta$ is $100\times$ larger in the enhanced 3D-FCN is that we did normalization in the enhanced loss 3D-FCN and $N_{neg}$ is much greater than $N_{pos}$.
We compare the last models in Table \ref{tab:3dfcn_elf} which shows the improvement of the enhanced loss function.

\subsection{Our 3D-FCN Implementation Details}
\label{ap:3dfcn}
The network details of 3D-FCN are shown in Table \ref{tab:network_details_3dfcn}.
Each Conv3D block in the BodyNet includes a 3D convolution layer, a ReLU layer and a batch normalization layer sequentially.
In the HeadNet, each Conv3D block represents an individual 3D convolution layer.
In the training phase, we create the ground truth for P-Map by setting the object-voxel which contains an object center as 1.
For the regression map, we create the ground truth by setting the object-voxels with 24-length residual vectors,
each of which is the coordinates for the eight points of the bounding box with a fixed order.
The result of the 3D-FCN baseline implemented by us is shown in the first row of Table \ref{tab:best_result}.

\subsection{Our VoxelNet Implementation Details}
The network details of VoxelNet are shown in Table \ref{tab:network_details_voxelnet}.
The FC block in VoxelNet consists of a fully connected layer, a batch normalization layer
and a ReLU layer sequentially.
Each Conv3D block in the MiddleLayer includes a 3D convolution layer, a ReLU layer and a batch normalization layer.
The Conv2D block in the RPN consists of a 2D convolution layer, a ReLU layer and a batch normalization layer.
The model of P-Map and R-Map is an individual 2D convolution layer.
We adopt the original parameterization method and residual vector for regression of VoxelNet\cite{Zhou_2018_CVPR}.
The result of our VoxelNet baseline is shown in the third row of Table \ref{tab:best_result}.

\begin{table}[tph]
    \begin{center}
    \caption[caption]{The Improvement of the Enhanced Loss Function for 3D-FCN}
    \label{tab:3dfcn_elf}
    \renewcommand\arraystretch{0.9}
    \renewcommand\tabcolsep{3pt}
    \begin{tabular}{@{}ccccccc@{}}
        \toprule
        \multirow{2}{*}{Detector}                          & \multicolumn{3}{c}{Bird's Eye View AP(\%)} & \multicolumn{3}{c}{3D Detection AP(\%)} \\
                                                                    & Easy                 & Mod                  & Hard                 & Easy                & Mod                 & Hard                \\ \midrule
        Original                                                    & 25.27                         & 21.48                         & 14.56                         & 15.45                        & 11.83                        & 12.13                        \\
        Enhanced                                                    & \textbf{28.90}                & \textbf{27.33}                & \textbf{27.47}                & \textbf{18.23}               & \textbf{16.54}               & \textbf{14.48}               \\ \bottomrule
        \end{tabular}
    \end{center}
    \vspace{-0.5cm}
  \end{table}
\begin{table}[tph]
    \caption{Our Implementation Details of 3D-FCN}
    \begin{center}
    \renewcommand\arraystretch{0.9}
    \renewcommand\tabcolsep{3pt}
    \begin{tabular}{@{}cccccc@{}}
        \toprule
        Block Name   & Layer Name & Kernel Size & Strides & Filter & GFLOPs \\ \midrule
        \multirow{4}{*}{Body} & conv3d\_1           & {[}5,5,5{]}          & {[}2,2,2{]}      & 32              & 25.8            \\
                              & conv3d\_2           & {[}5,5,5{]}          & {[}2,2,2{]}      & 64              & 204.9           \\
                              & conv3d\_3           & {[}3,3,3{]}          & {[}2,2,2{]}      & 96              & 16.6            \\
                              & conv3d\_4           & {[}3,3,3{]}          & {[}1,1,1{]}      & 96              & 24.9            \\ \midrule
        Head-PMap             & conv3d\_obj         & {[}3,3,3{]}          & {[}1,1,1{]}      & 1               & 0.3             \\
        Head-RMap             & conv3d\_cor         & {[}3,3,3{]}          & {[}1,1,1{]}      & 24              & 6.2             \\ \bottomrule
        \end{tabular}
    \end{center}
    \vspace{-0.5cm}
    \label{tab:network_details_3dfcn}
\end{table}
\begin{table}[tph]
    \caption{Our Implementation Details of VoxelNet}
    \begin{center}
    \renewcommand\arraystretch{0.9}
    \renewcommand\tabcolsep{3pt}
    \begin{tabular}{@{}cccccc@{}}
        \toprule
        Block Name          & Layer Name & \begin{tabular}[c]{@{}c@{}}Kernel Size /\\ Output Unit\end{tabular} & Strides & Filter & GFLOPs \\ \midrule
        \multirow{3}{*}{FeatureNet}  & vfe                 & 32                                                                           & N/A              & N/A             & \textless{}0.1  \\
                                     & vfe                 & 128                                                                          & N/A              & N/A             & \textless{}0.1  \\
                                     & fc                  & 128                                                                          & N/A              & N/A             & \textless{}0.1  \\ \midrule
        \multirow{4}{*}{MiddleLayer} & conv3d              & {[}3,3,3{]}                                                                  & {[}2,1,1{]}      & 64              & 311.5           \\
                                     & conv3d              & {[}3,3,3{]}                                                                  & {[}1,1,1{]}      & 64              & 93.5            \\
                                     & conv3d              & {[}3,3,3{]}                                                                  & {[}2,1,1{]}      & 64              & 62.3            \\
                                     & reshape             & N/A                                                                          & N/A              & N/A             & /               \\ \midrule
        \multirow{9}{*}{RPN}         & conv2d              & {[}3,3{]}                                                                    & {[}2,2{]}        & 128             & 41.6            \\
                                     & conv2d$\times$3            & {[}3,3{]}                                                                    & {[}1,1{]}        & 128             & 31.2            \\
                                     & deconv              & {[}3,3{]}                                                                    & {[}1,1{]}        & 256             & 20.8            \\
                                     & conv2d              & {[}3,3{]}                                                                    & {[}2,2{]}        & 128             & 10.4            \\
                                     & conv2d$\times$5            & {[}3,3{]}                                                                    & {[}1,1{]}        & 128             & 13.0            \\
                                     & deconv              & {[}2,2{]}                                                                    & {[}2,2{]}        & 256             & 5.2             \\
                                     & conv2d              & {[}3,3{]}                                                                    & {[}2,2{]}        & 256             & 5.2             \\
                                     & conv2d$\times$5            & {[}3,3{]}                                                                    & {[}1,1{]}        & 256             & 13.0            \\
                                     & deconv              & {[}4,4{]}                                                                    & {[}4,4{]}        & 256             & 2.6             \\ \midrule
        Prob-Map                     & conv2d              & {[}1,1{]}                                                                    & {[}1,1{]}        & 2               & 0.1             \\
        Reg-Map                      & conv2d              & {[}1,1{]}                                                                    & {[}1,1{]}        & 14              & 0.8             \\ \bottomrule
        \end{tabular}
        \end{center}
        \label{tab:network_details_voxelnet}
\end{table}

